\documentclass[10pt,twocolumn,letterpaper]{article}

\usepackage{iccv}
\usepackage{times}
\usepackage{epsfig}
\usepackage{graphicx}
\usepackage{amsmath}
\usepackage{amssymb}
\usepackage{amsthm}
\usepackage{dblfloatfix}
\usepackage{amsmath,amsfonts,bm}

\def\eqref#1{equation~\ref{#1}}
\def\1{\bm{1}}

\def\mB{{\bm{B}}}
\def\mC{{\bm{C}}}

\def\mH{{\bm{H}}}

\def\mM{{\bm{M}}}

\def\mX{{\bm{X}}}
\def\mY{{\bm{Y}}}

\DeclareMathAlphabet{\mathsfit}{\encodingdefault}{\sfdefault}{m}{sl}
\SetMathAlphabet{\mathsfit}{bold}{\encodingdefault}{\sfdefault}{bx}{n}

\def\sI{{\mathbb{I}}}

\newtheorem{definition}{Definition}
\newtheorem{proposition}{Proposition}
\usepackage[pagebackref=true,breaklinks=true,letterpaper=true,colorlinks,bookmarks=false]{hyperref}

\iccvfinalcopy % *** Uncomment this line for the final submission

\def\iccvPaperID{1707} % *** Enter the ICCV Paper ID here

\ificcvfinal\fi
\begin{document}

%%%%%%%%% TITLE
\title{Interpretable Spatio-temporal Attention for Video Action Recognition}

\author{Lili Meng, Bo Zhao, Bo Chang\\
University of British Columbia\\
{\tt\small \{menglili@cs, bzhao03@cs, bchang@stat\}.ubc.ca}
\and
Gao Huang\\
Tsinghua University\\
{\tt\small gaohuang@tsinghua.edu.cn}
\and
Wei Sun, Frederich Tung, Leonid Sigal\\
University of British Columbia\\
{\tt\small \{wei.sun@alumni, ftung@cs, lsigal@cs\}.ubc.ca}
}

\maketitle

\begin{abstract}
  Inspired by the observation that humans are able to process videos efficiently by only paying attention \emph{where} and \emph{when} it is needed, we propose an interpretable and easy plug-in spatial-temporal attention mechanism for video action recognition. For spatial attention, we learn a saliency mask to allow the model to focus on the most salient parts of the feature maps. For temporal attention, we employ a convolutional LSTM based attention mechanism to identify the most relevant frames from an input video. Further, we propose a set of regularizers to ensure that our attention mechanism attends to coherent regions in space and time.  Our model not only improves video action recognition accuracy, but also localizes discriminative regions both spatially and temporally, despite being trained in a weakly-supervised manner with only classification labels (no bounding box labels or time frame temporal labels). We evaluate our approach on several public video action recognition datasets with ablation studies. Furthermore, we quantitatively and qualitatively evaluate our model's ability to localize discriminative regions spatially and critical frames temporally. Experimental results demonstrate the efficacy of our approach, showing superior or comparable accuracy with the state-of-the-art methods while increasing model interpretability. 
  
\end{abstract}
\section{Introduction} 

%\begin{figure}[t]
 %   \begin{center}
  %      \includegraphics[width = \linewidth]{figures/spatial_temporal_example.pdf} 
  %  \end{center}
 %   \caption{\textbf{Spatio-temporal attention for video action recognition.} The first row is the input frames, and the second row demonstrates the spatial-temporal attention map of each frame. The redder region represents the higher spatial importance and the globally brighter frame means it is more temporally important.}
   % \label{fig:spatial_temporal_example}
%\end{figure}

An important property of human perception is that one does not need to process a scene in its entirety at once. Instead, humans focus attention selectively on parts of the visual space to acquire information \emph{where} and \emph{when} it is needed, and combine information from different fixations over time to build up an internal representation of the scene~\cite{rensink2000dynamic}, which can be used for interpretation or prediction.

In computer vision and natural language processing, over the last couple of years, attention models have proved important, particularly for tasks where interpretation or explanation requires only a small portion of the image, video or sentence. Examples include visual question answering \cite{lu2016hierarchical,xiong2016dynamic,xu2016ask}, activity recognition \cite{girdhar2017attentional,li2018videolstm,sharma2015action}, and neural machine translation \cite{bahdanau2014neural, vaswani2017attention}. These models have also provided certain interpretability, by visualizing regions selected or attended over for a particular task or decision. In particular, for video action recognition, a proper attention model can help answer the question of not only “\emph{where} but also \emph{when} it needs to look” at the image evidence to draw a decision. It intuitively explains which part the model attends to and provides easy-to-interpret rationales when making a particular decision. Such interpretability is helpful and even necessary for some real applications, \eg, medical AI systems~\cite{zhangetal2017mdnet} or self-driving cars~\cite{kim2017interpretable}.

\begin{figure*}[t]
    \begin{center}
        \includegraphics[width = 0.98\linewidth]{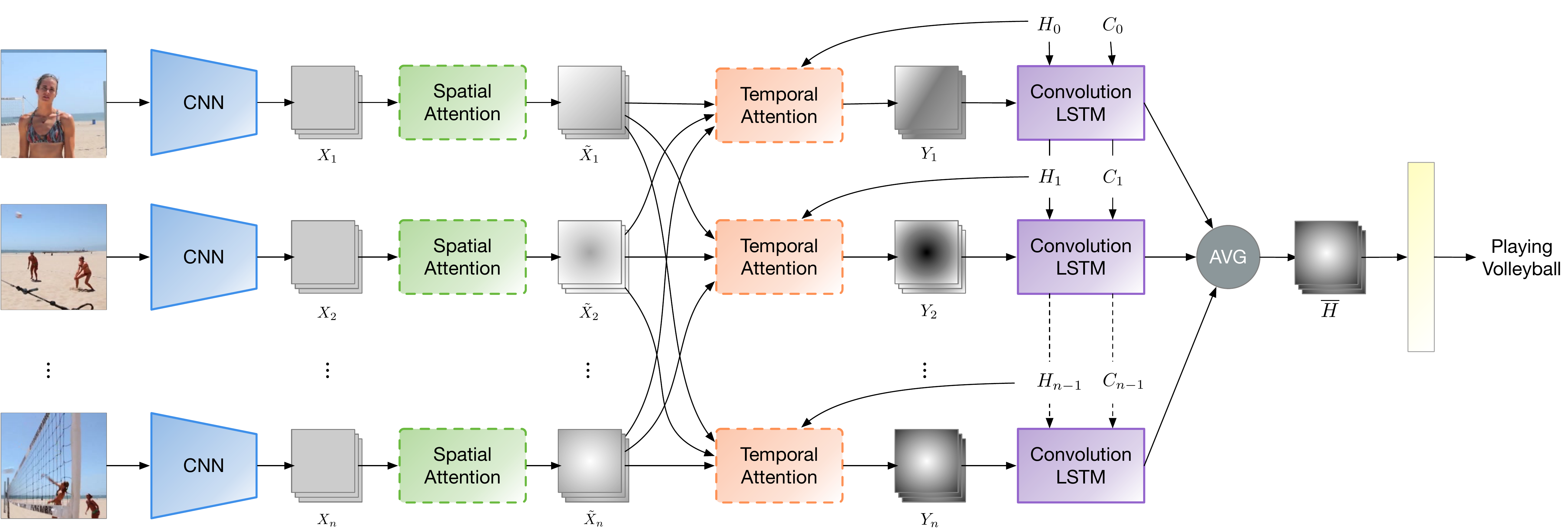} 
    \end{center}
    \caption{\textbf{Spatio-temporal attention for video action recognition.} The convolutional features are attended over both spatially, in each frame, and subsequently temporally. Both attentions are soft, meaning that the effective final representation at time $t$ of an RNN, used to make the prediction, is a spatio-temporally weighted aggregation of convolutional features across the video along with the past hidden state from $t-1$. For details please refer to Sec. \ref{sec:method}.
    }
    \label{fig:spatial_temporal_att}
\end{figure*}

Visual attention for action recognition in videos is challenging as videos contain both spatial and temporal information. 
Previous work in attention-based video action recognition either focused on spatial attention \cite{girdhar2017attentional,li2018videolstm,sharma2015action} or temporal attention \cite{torabi2017action}. Intuitively, both spatial attention and temporal attention should be integrated together to help make and explain the final action prediction.
In this paper, we propose a novel spatio-temporal attention mechanism that is designed to address these challenges. Our attention mechanism is efficient, due to it’s space- and time- separability, and yet flexible enough to enable encoding of effective regularizers (or priors). As such, our attention mechanism consists of spatial and temporal components shown in Fig.~\ref{fig:spatial_temporal_att}. The spatial attention component, which attenuates frame-wise CNN image features, consists of the saliency mask, regularized to be discriminative and spatially smooth. The temporal component consists of a uni-modal soft attention mechanism that aggregates information over the nearby attenuated frame features before passing it into a convolutional LSTM for class prediction. 

\vspace{0.1in}
\noindent
{\bf Contributions:} In summary, the main contributions of this work are: 
(1) We introduce an interpretable and easy plug-in spatial-temporal attention for video action recognition, which consists of the saliency mask for spatial attention learned by ConvNets and temporal attention learned by convolutional LSTM. (2) We introduce three different regularizers, two for spatial and one for temporal attention components, to improve performance and interpretability of our model; (3) We demonstrate the efficacy of our model for video action recognition on three public datasets and explore the importance of our modeling choices through ablation experiments;
(4) Finally, we qualitatively and quantitatively show that our spatio-temporal attention is able to localize discriminative regions and important frames, despite being trained in a purely weakly-supervised manner with only classification labels.

\section{Related work}
\subsection{Visual explanations of neural networks}
Various methods have been proposed to provide explanations of neural networks by visualization \cite{mahendran2016salient,ramprasaath2016grad,simonyan2013deep,springenberg2014striving,zeiler2014visualizing,zhang2016top, zhang2018interpretable}, including visualizing the gradients, perturbing the inputs, and bridging relations with other well-studied systems. The class activation mapping (CAM) applies the global average pooling layer for identifying discriminative regions \cite{zhou2016learning}.
The gradient-weighted class activation mapping (Grad-CAM) utilizes the gradient flow and relaxes the architectural assumptions made by CAM \cite{selvaraju2017grad}.
The guided attention inference network (GAIN)
makes the network's attention trainable
in an end-to-end fashion \cite{li2018tell}. Visual attention is one way to explain which part of the image is responsible for the network's decision \cite{jetley2018learn, kim2017interpretable}.  A visual attention model is used to train a ConvNet regressor from images to steering angle for self-driving cars, and image regions that potentially influence the network's output are highlighted by the model \cite{kim2017interpretable}. A semantically and visually interpretable medical image diagnosis network is proposed to explore discriminative image feature descriptions from reports ~\cite{zhangetal2017mdnet}. However, few work focus on visual interpretation both spatially and temporally. 

\subsection{Visual attention for video action recognition}
Video action recognition is one of the fundamental problems in computer vision and has been widely studied \cite{carreira2017quo, feichtenhofer2016convolutional, hara2018can, tran2015learning, simonyan2014two, tran2018closer}. A recent survey can be found in \cite{kong2018human}, so here we only focus on attention-based models \cite{du2018recurrent, girdhar2017attentional, li2018videolstm, sharma2015action, song2017end, torabi2017action}. For video action recognition, visualizing which part of the frame and which frame of the video sequence that the model was attending to provides valuable insight into the model's behavior. An attention-driven LSTM \cite{sharma2015action} is proposed for action recognition and it can highlight important spatial locations. Attentional pooling \cite{girdhar2017attentional} introduces an attention mechanism based on a derivation of bottom-up and top-down attention as low-rank approximations of bilinear pooling methods. However, these work only focus on the crucial spatial locations of each image, without considering temporal relations among different frames in a video sequence. To alleviate this shortcoming, visual attention is incorporated in the motion stream \cite{du2018recurrent, li2018videolstm, wang2016hierarchical}. However, the motion stream only employs the optical flow frames generated from consecutive frames, and cannot consider the long-term temporal relations among different frames in a video sequence. Moreover, motion stream needs additional optical flow frames as input, which imposes burden due to additional optical flow extraction, storage and computation and is especially severe for large datasets.  \cite{torabi2017action} proposes an attention based LSTM model to highlight frames in videos, but spatial information is not used for temporal attention. An end-to-end spatial and temporal attention model is proposed in \cite{song2017end} for human action recognition, but additional skeleton data is needed.

\subsection{Spatial and temporal action localization}
Spatial and temporal action localization have been long studied and here we constrain the discussion to some recent learning based methods \cite{shou2016action,singh2017online,wang2017untrimmednets, weinzaepfel2015learning,xu2017r,  yeung2016end, yu2015fast}. For spatial localization, \cite{weinzaepfel2015learning} proposes to detect proposals at the frame-level and scores them with a combination of static and motion CNN features, and then tracks high-scoring proposals throughout the video using a tracking-by-detection approach. However, these models are trained in a supervised manner in which bounding box annotations are required. For temporal localization, \cite{yeung2016end} employs REINFORCE to learn the agent's decision policy for action detection. UntrimmedNet \cite{wang2017untrimmednets} couples the classification and the selection module to learn the action models and reason about the action temporal duration in a weakly supervised manner. However, the above mentioned methods consider spatial and temporal localization separately. A spatio-temporal action localization is proposed to localize both spatially and temporally in \cite{singh2017online}, but both spatial and temporal labels are required. It is very expensive and time-consuming to acquire these per-frame labels in large-scale video datasets.

Our proposed method can localize discriminative regions both spatially and temporally besides improving action recognition accuracy, despite being trained in a weakly-supervised fashion with only classification labels (no bounding box annotations or time frame temporal labels).
\section{Spatial-temporal attention mechanism}
\label{sec:method}
Our overall model is an Recurrent Neural Network (RNN) that aggregates frame-based convolutional features across the video to make action predictions as shown in Fig.~\ref{fig:spatial_temporal_att}. The convolutional features are attended over both spatially, in each frame, and subsequently temporally for the entire video sequence. Both attentions are soft, meaning that the effective final representation at time $t$ is a spatio-temporally weighted aggregation of convolutional features across the video along with the past hidden state from $t-1$. The core novelty is the overall form of our spatio-temporal attention mechanism and the additional terms of the loss function that induce sensible spatial and temporal attention priors. 

\subsection{Convolutional frame features}
We use the 
% The features used in this work is the 
last convolutional layer output extracted by ResNet50 or ResNet101 \cite{he2016deep}, pretrained on the ImageNet \cite{deng2009imagenet} dataset and fine-tuned for the target dataset, as our frame feature representation. 
%As a result, out feature vector $X$ is of dimension \fix{\textcolor{red}{FILL IN}}. 
We acknowledge that more accurate feature extractors (for instance, network with more parameters such as ResNet-152 or higher performance networks such as DenseNet \cite{huang2017densely} or SENet \cite{hu2017squeeze}) and optical flow features will likely lead to better overall performance. Our primary purpose in this paper is to prove the efficacy of our spatial-temporal attention mechanism. Hence we keep the features relatively simple.

\subsection{Spatial attention with importance mask}
\label{subsec: spatial_attention}
\begin{figure}
    \begin{center}
        \includegraphics[width = 0.95\linewidth]{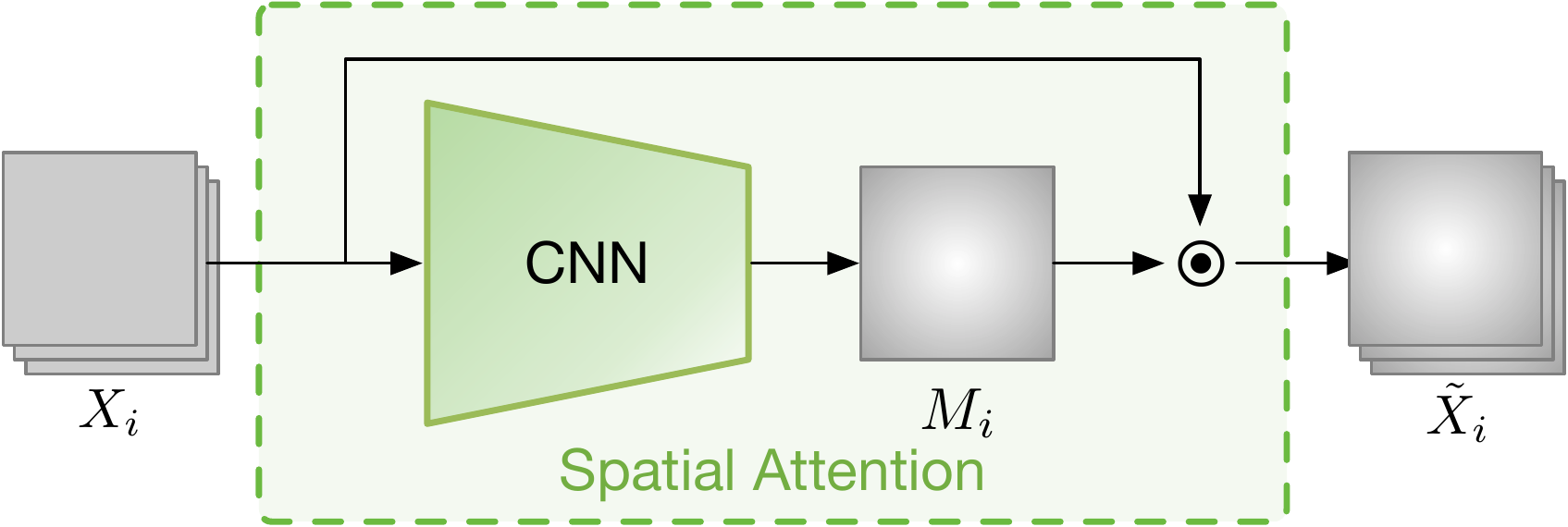} 
    \end{center}
    \caption{\textbf{Spatial attention component.} We use several layers of convolutional network to learn the importance mask $\mM_i$ for the input image feature $\mX_i$, the output is the element-wise multiplication $\tilde{\mX}_i = \mX_i \odot \mM_i$. Details please refer to Sec. \ref{subsec: spatial_attention}.
    } 
    \label{fig:spatial_attention}
\end{figure}
We apply an \emph{importance mask} $\mM_i$ to the image feature $\mX_i$ of the $i$-th frame to obtain attended image features by element-wise multiplication:
\begin{equation}
\tilde{\mX}_i = \mX_i \odot \mM_i,
\end{equation}
for $1 \leq i \leq n$, where $n$ is the number of frames, and each entry of $\mM$ lies in $[0,1]$.
This operation attenuates certain regions of the feature map based on their estimated importance. Here we simply use three convolutional layers to learn the importance mask. Fig. \ref{fig:spatial_attention} illustrates our spatial attention component and Table \ref{tb:spatial_attention} shows the network architecture details. 
However, if left unconstrained, an arbitrarily structured mask could be learned, leading to possible overfitting. We posit that, in practice, it is often useful to attend to a few important larger regions (\eg, objects, elements of the scene). To induce this behavior, we encourage smoothness of the mask by introducing total variation loss on the spatial attention, as will be described in Sec. \ref{subsec:loss_function}.

\begin{table}[b]
\center
\small
\begin{tabular}{c|c}
\hline
\textbf{Index} & \textbf{Operation} \\ \hline
(1) & $\mathbf{X}_i$  \\
(2) &  CONV-(N1024, K3, S1, P1), BN, ReLU \\
(3) & CONV-(N512, K3, S1, P1), BN, ReLU \\
(4) &  CONV-(N1, K3, S1, P1), Sigmoid 
\\\hline
\end{tabular}
\vspace{+3mm}
\caption{\textbf{Architecture of our spatial attention module} (N: number of channels, K: kernel size, S: stride, P: padding). }
\label{tb:spatial_attention}
\end{table}

\subsection{Temporal attention based on ConvLSTM} 
\label{subsec: temporal_attention_methods}
We introduce the temporal attention mechanism inspired by attention for neural machine translation~\cite{bahdanau2014neural}. However, conventional LSTM in neural machine translation uses full connections in the input-to-state and state-to-state transitions and cannot encode the spatial information in images. Therefore, to mitigate this drawback, we use Convolutional LSTM (ConvLSTM) \cite{xingjian2015convolutional} instead. For ConvLSTM, each input, cell output, hidden state, and gate are 3D tensors whose last two dimensions are spatial dimensions. These 3D tensors can preserve spatial information, which is more suitable for image inputs.

Our temporal attention mechanism generates {\em energy} for each attended frame $\tilde{\mX}_i$ at each time step $t$, 
\begin{equation}
e_{ti} = \Phi (\mH_{t-1}, \tilde{\mX}_i), 
\end{equation}
where $\mH_{t-1}$ represents the ConvLSTM hidden state at time $t-1$ that implicitly contains all previous information up to time step $t-1$, $\tilde{\mX}_i$ represents the $i$-th frame masked features, and $\Phi(\mH_{t-1},\tilde{\mX}_i) = \Phi_H(\mH_{t-1})+\Phi_X(\tilde{\mX}_i)$, where $\Phi_H$ and $\Phi_X$ are feed-forward neural networks which are jointly trained with all other components of the proposed system. 

\begin{figure}[t]
    \begin{center}
        \includegraphics[width = \linewidth]{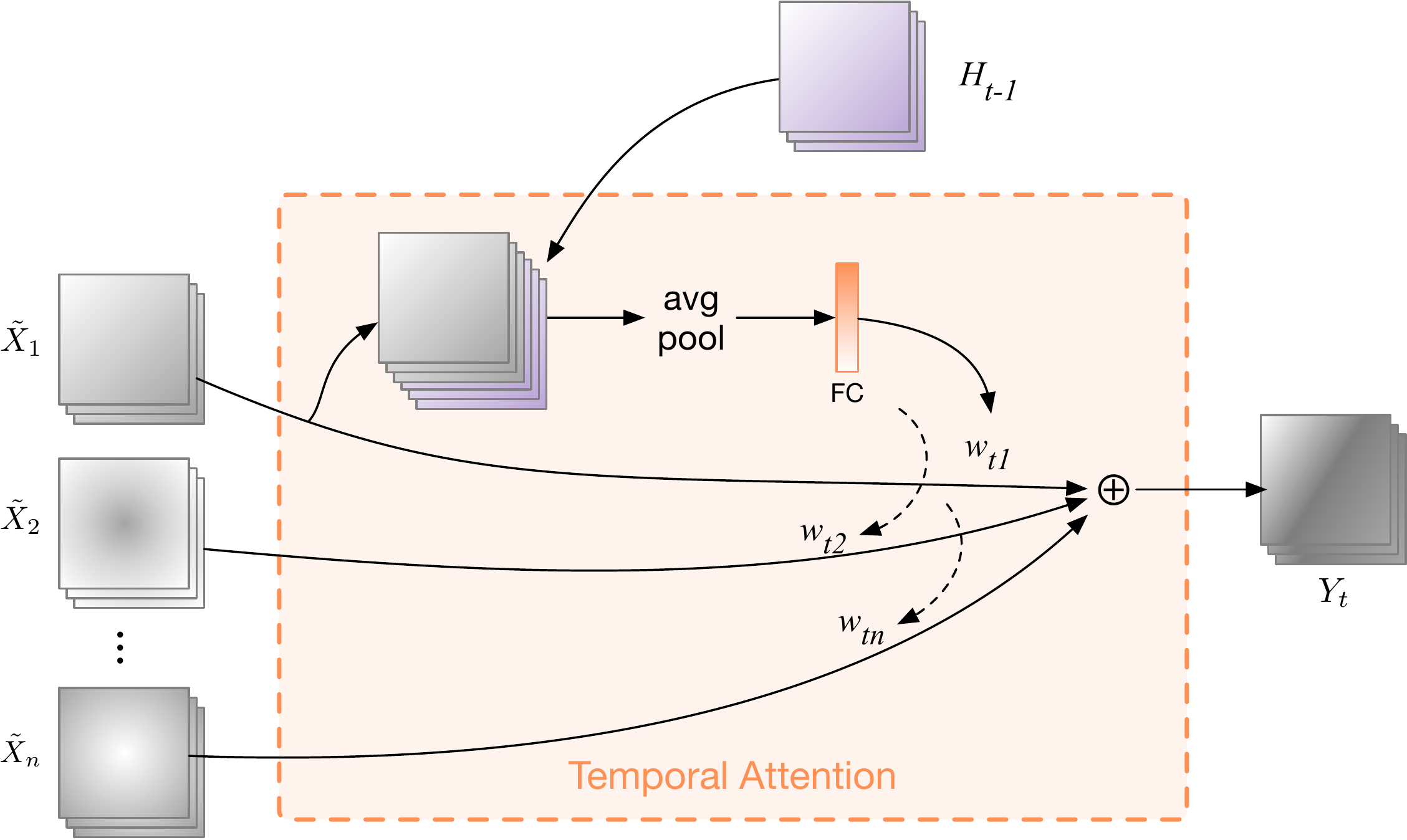} 
    \end{center}
    \caption{\textbf{Temporal attention component.} The temporal attention learns a temporal attention weight $w_{ti}$ at each time step $t$. The final feature map $\mY_t$ at time $t$ to the ConvLSTM is a weighted sum of the feature from all the previous masked frames. For details please refer to Sec. \ref{subsec: temporal_attention_methods}.
    }
    \label{fig:temporal_attention_illustration}
    \vspace{-3mm}
\end{figure}

This temporal attention model directly computes a soft attention weight for each frame at each time $t$ as shown in Fig.~\ref{fig:temporal_attention_illustration}. It allows the gradient of the cost function to be backpropagated through. This gradient can be used to train the entire spatial-temporal attention model jointly.

The importance weight $w_{ti}$ for each frame is:
\begin{equation}
w_{ti}=\frac{\exp(e_{ti})}{\sum_{i=1}^{n}(\exp(e_{ti}))},
\end{equation}
for $1 \leq i \leq n, 1 \leq t \leq n$.
This importance weighting mechanism decides which frame of the video to pay attention to.
The final feature map $\mY_t$ that is input to the ConvLSTM is a weighted sum of the features from all of the frames:
\begin{equation}
\mY_t = \frac{1}{n}\sum_{i=1}^{n} w_{ti} \tilde{\mX}_i,
\end{equation}
where $\tilde{\mX}_i$ denotes the $i$-th masked frame of each video. %, $n$ represents the total number of frames for each video.

%In contrast, in ConvLSTM, each input $\mY_{t}$, cell output $\mO_t$, hidden state $\mH_t$ and gate $\mI_t, \mF_t, \mO_t$ are 3D tensors whose last two dimensions are spatial dimensions (rows and columns).
%\begin{align*}
%\mI_t &= \sigma (\mW_{xi} * \mY_{t} + \mW_{hi}*\mH_{t-1}+
% W_{ci}\circ C_{t-1}+
%\vb_i),\\
%\mF_t &= \sigma (\mW_{xf} * \mY_{t} + \mW_{hf}*\mH_{t-1}+
% W_{ci}\circ C_{t-1}+
%\vb_f),\\
%\mC_t &=\mF_t \circ \mC_{t-1} + \mI_t\circ \tanh(\mW_{xc}* %\mY_{t}+\mW_{hc}*\mH_{t-1}+\vb_c),\\
%\mO_t &= \sigma (\mW_{xo}*\mY_{t} + \mW_{ho}*\mH_{t-1}+
% W_{co}\circ C_{t-1}+
%\vb_o),\\
%\mH_t &= \mO_t \circ \tanh(\mC_t).
%\end{align*}

We use the following initialization strategy for the ConvLSTM cell state and hidden state for faster convergence:
\begin{equation}
\mC_0 = g_{c}(\frac{1}{n}\sum_{i=1}^n \tilde{\mX}_i), \quad \quad \mH_0 = g_{h}(\frac{1}{n}\sum_{i=1}^n \tilde{\mX}_i),
\end{equation}
where $g_{c}$ and $g_{h}$ are two layer convolutional networks with batch normalization \cite{ioffe2015batch}.

We calculate the average hidden states of ConvLSTM over time length $n$,
% \begin{equation}
$
    \overline{\mH} = \frac{1}{n} \sum_{i=1}^n \mH_i
$
% \end{equation}
and send it to a fully connected classification layer for the final video action classification.

\subsection{Loss function}
\label{subsec:loss_function}
Considering the spatial and temporal nature of our video action recognition, we would like to learn (1) a sensible attention mask for spatial attention, (2) reasonable importance weighting scores for different frames, and (3) improve the action recognition accuracy at the same time. Therefore, we define our loss function $L$ as:
\begin{equation}
L = L_{\mathrm{CE}} + \lambda_{\mathrm{TV}} L_{\mathrm{TV}} +  \lambda_{\mathrm{contrast}} L_{\mathrm{contrast}}  + \lambda_{\mathrm{unimodal}} L_{\mathrm{unimodal}},
\end{equation}
where $L_{\mathrm{CE}}$ is the cross entropy loss for classification, 
% $L_{\mathrm{TV}}$ and $L_{\mathrm{contrast}}$ are regularizers on spatial attention masks and $L_{\mathrm{unimodal}}$ is the regularizer on the temporal soft attention. More specifically, 
$L_{\mathrm{TV}}$ represents the total variation regularization~\cite{rudin1992nonlinear}; $L_{\mathrm{contrast}}$ represents the mask and background contrast regularizer; and $L_{\mathrm{unimodal}}$ represents unimodality regularizer. 
$\lambda_{\mathrm{TV}}$, $\lambda_{\mathrm{contrast}}$ and $\lambda_{\mathrm{unimodal}}$ are the weights for the corresponding regularizers.

The total variation regularization $L_{\mathrm{TV}}$ of the learnable attention mask encourages spatial smoothness of the mask and is defined as:
\begin{equation}
L_{\mathrm{TV}} \!=\! \sum_{i=1}^n 
\!\left(\!
\sum_{j, k}|\mM_i^{j+1,k}\!-\!\mM_i^{j,k}|\!+\!\sum_{j,k}|\mM_i^{j,k+1}\!-\!\mM_i^{j,k}|
\!\right)
\end{equation}
where $\mM_i$ is the mask for the $i$-th frame, and $\mM_i^{j,k}$ is the entry at the $(j, k)$-th spatial location of the mask. 
The contrast regularization $L_{\mathrm{contrast}}$ of learnable attention mask is used to suppress the irrelevant information and highlight important parts:
\begin{equation}
L_{\mathrm{contrast}} = \sum_{i=1}^n 
\left(
-\frac{1}{2} \mM_i \odot \mB_i + \frac{1}{2} \mM_i \odot (1-\mB_i)
\right)
\end{equation}
where $\mB_i = \sI \{ \mM_i > 0.5 \}$ represents the binarized mask,
% , threshold value larger than $0.5$. 
$\sI$ is an indicator function applied element-wise.
%\BC{Add citation for $L_{\mathrm{contrast}}$.}

The unimodality regularizer $L_{\mathrm{unimodal}}$ encourages the temporal attention weights to be unimodal, biasing against spurious temporal weights. This stems from our observation that in most cases only one activity would be present in the considered frame window, with possible irrelevant information on either or both sides. Here we use the log concave distribution to encourage the unimodal pattern of temporal attention weights:
\begin{equation}
L_{\mathrm{unimodal}} = 
\sum_{t=1}^n \sum_{i=2}^{n-1}
\max\{0,
w_{t,i-1} w_{t,i+1} - w_{t,i}^2
\}
\end{equation}
%where $T$ represents the ConvLSTM time sequence length and $n$ is the number of frames for each video. 
%For more details on this log concave sequence we will introduce in next Section \ref{sec: log_concave}.

\subsection{Log-concave regularization}
\label{sec: log_concave}

A probability distribution is unimodal if it has a single peak or mode.
The temporal attention weights are a univariate discrete distribution over the frames, indicating the importance of the frames for the task of classification.
In the context of activity recognition, it is reasonable to assume that the frames that contain salient information should be consecutive, instead of scattered around. 
Therefore, we introduce a mathematical concept called the log-concave sequence and design a regularizer that encourages unimodality.
We first give a formal definition of the unimodal sequence.
\begin{definition}
A sequence $\{a_i\}_{i=1}^n$ is unimodal if for some integer m,
\[
\begin{cases}
a_{i-1} \leq  a_i &\quad \mathrm{if} \ i \leq m, \\
a_i \geq a_{i+1} &\quad \mathrm{if} \ i \geq m.
\end{cases}
\]
\end{definition}
A univariate discrete distribution is unimodal, if its probability mass function forms a unimodal sequence.
The log-concave sequence is defined as follows.
\begin{definition}
A non-negative sequence $\{a_i\}_{i=1}^n$ is log-concave if $a_i^2 \geq a_{i-1} a_{i+1}$.
\end{definition}
This property gets its name from the fact that if $\{a_i\}_{i=1}^n$ is log-concave, then the sequence $\{\log a_k\}_{i=1}^n$ is concave.
The connection between unimodality and log-concavity is given by the following proposition.
\begin{proposition}
    \label{prop:log_cancave}
    A log-concave sequence is unimodal.
\end{proposition}
The proof of this proposition is given in the supplementary material. Given the definition of log-concavity, it is straightforward to design a regularization term that encourages log-concavity:
\begin{equation}
    R = \sum_{i=2}^{n-1} \max\{0, a_{i-1} a_{i+1} - a_i^2\}.
\end{equation}
By Proposition~\ref{prop:log_cancave}, this regularizer also encourages unimodality.
\section{Experiments}
In this section, we first conduct experiments to evaluate our proposed method on video action recognition tasks on three publicly available datasets. We then evaluate our spatial attention mechanism on the spatial localization task and our temporal attention mechanism on the temporal localization task respectively. 
\subsection{Video action recognition}
We first conduct extensive studies on the widely used HMDB51~\cite{kuehne2011hmdb} and UCF101~\cite{soomro2012ucf101} datasets. The purpose of these experiments is mainly to examine the effects of different sub-components. We then show that our method can be applied to the challenging large-scale Moments in Time dataset~\cite{monfort2018moments}.

\begin{figure*}[!tbh]
\centering

\begin{minipage}{0.135\textwidth}
\centering
\includegraphics[width=\textwidth]{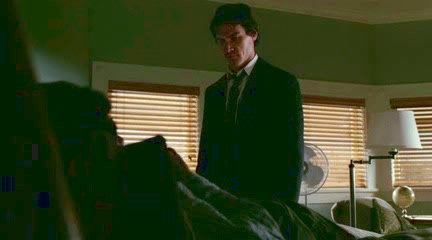}
\end{minipage} 
\begin{minipage}{0.135\textwidth}
\centering
\includegraphics[width=\textwidth]{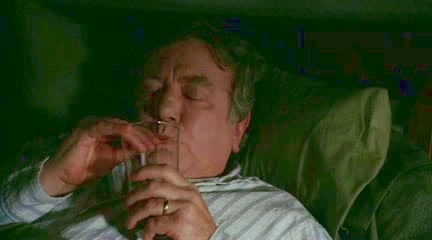}
\end{minipage}
\begin{minipage}{0.135\textwidth}
\centering
\includegraphics[width=\textwidth]{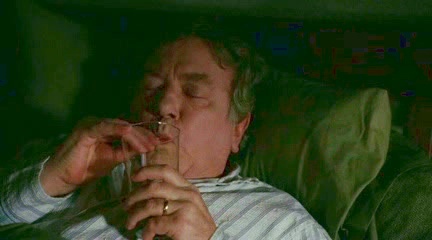}
\end{minipage}
\begin{minipage}{0.135\textwidth}
\centering
\includegraphics[width=\textwidth]{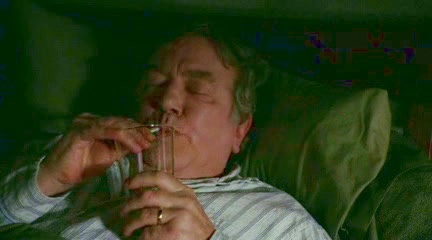}
\end{minipage} 
\begin{minipage}{0.135\textwidth}
\centering
\includegraphics[width=\textwidth]{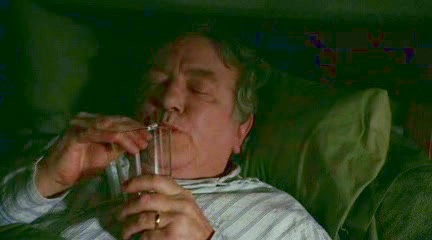}
\end{minipage} 
\begin{minipage}{0.135\textwidth}
\centering
\includegraphics[width=\textwidth]{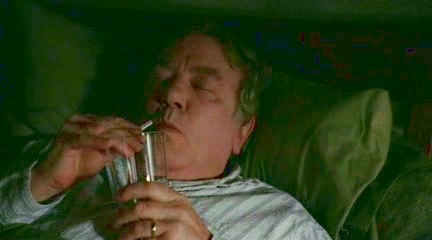}
\end{minipage} 
\begin{minipage}{0.135\textwidth}
\centering
\includegraphics[width=\textwidth]{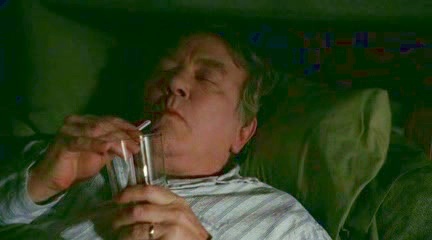}
\end{minipage} 
\begin{minipage}{0.135\textwidth}
\centering
\includegraphics[width=\textwidth]{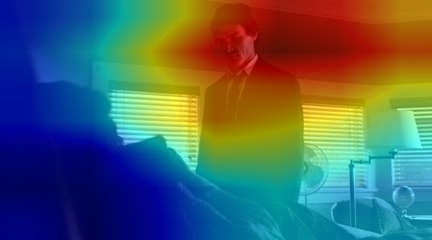}
\end{minipage} 
\begin{minipage}{0.135\textwidth}
\centering
\includegraphics[width=\textwidth]{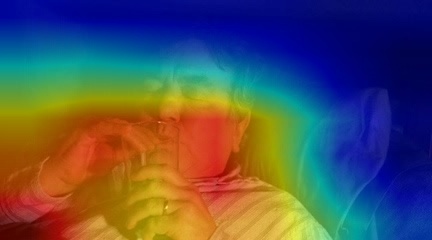}
\end{minipage}
\begin{minipage}{0.135\textwidth}
\centering
\includegraphics[width=\textwidth]{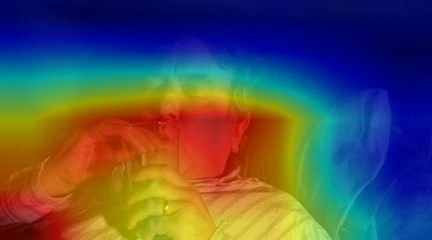}
\end{minipage}
\begin{minipage}{0.135\textwidth}
\centering
\includegraphics[width=\textwidth]{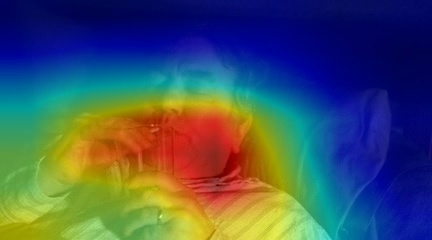}
\end{minipage} 
\begin{minipage}{0.135\textwidth}
\centering
\includegraphics[width=\textwidth]{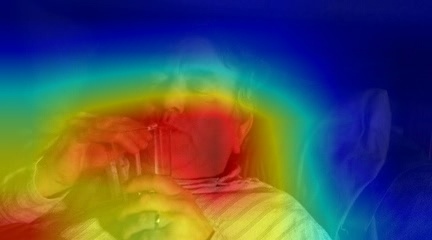}
\end{minipage} 
\begin{minipage}{0.135\textwidth}
\centering
\includegraphics[width=\textwidth]{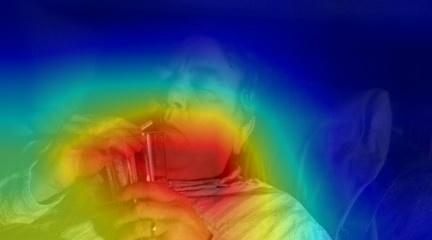}
\end{minipage} 
\begin{minipage}{0.135\textwidth}
\centering
\includegraphics[width=\textwidth]{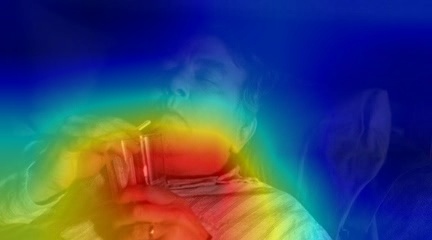}
\end{minipage} 
\begin{minipage}{0.135\textwidth}
\centering
\includegraphics[width=\textwidth]{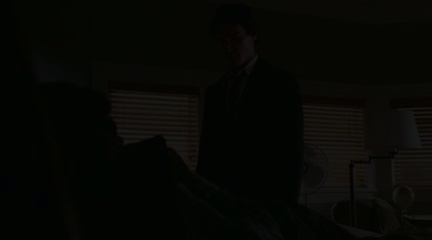}
{(a) frame01}
\end{minipage} 
\begin{minipage}{0.135\textwidth}
\centering
\includegraphics[width=\textwidth]{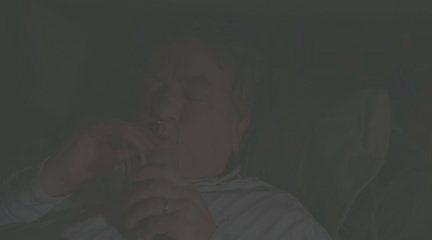}
{(b) frame11}
\end{minipage}
\begin{minipage}{0.135\textwidth}
\centering
\includegraphics[width=\textwidth]{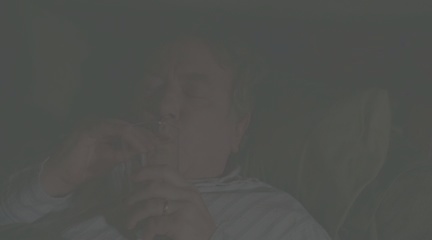}
{(c) frame29}
\end{minipage}
\begin{minipage}{0.135\textwidth}
\centering
\includegraphics[width=\textwidth]{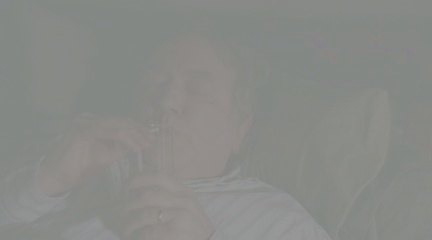}
{(d) frame55}
\end{minipage} 
\begin{minipage}{0.135\textwidth}
\centering
\includegraphics[width=\textwidth]{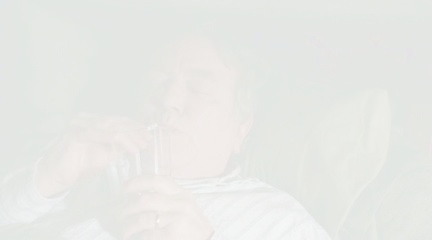}
{(e) frame59}
\end{minipage} 
\begin{minipage}{0.135\textwidth}
\centering
\includegraphics[width=\textwidth]{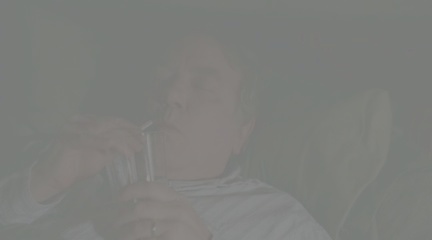}
{(f) frame67}
\end{minipage} 
\begin{minipage}{0.135\textwidth}
\centering
\includegraphics[width=\textwidth]{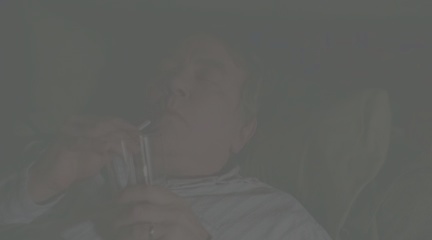}
{(f) frame89}
\end{minipage} 
\vspace{+3mm}
\caption{\textbf{Examples of spatial temporal attention.} (Best viewed in color.) A frame sequence from a video of \emph{Drink} action in HMDB51. The original images are shown in the top row, spatial attention is shown as heatmap (red means important) in the middle row, and temporal attention score is shown as the gray image (the brighter the frame is, the more crucial the frame is) in the bottom row. It shows that spatial attention can focus on important areas while temporal attention can attend to crucial frames. The temporal attention also shows a unimodal distribution for the entire action from starting to drink to completing the action.}
\label{fig:attention_hmdb51}
\end{figure*}  

\vspace{0.07in}
\noindent
\textbf{Datasets.} HMDB51 dataset \cite{kuehne2011hmdb} contains 51 distinct action categories, each containing at least 101 clips for a total of 6,766 video clips extracted from a wide range of sources. These videos include general facial actions, general body movements, body movements with object interaction, and body movements for human interaction. 

UCF101 dataset \cite{soomro2012ucf101} is an action recognition dataset of realistic action videos, collected from YouTube, with 101 action categories. 

Moments in Time Dataset \cite{monfort2018moments} is a collection of one million short videos with one action label per video and 339 different action classes.  %As there could be more than one action taking place even in a video , which may hinder the performance of action recognition models which 
As there could be more than one action taking place in a video, action recognition models may predict an action correctly yet be penalized because the ground truth does not include that action. Therefore, it is believed that top 5 accuracy measure will be more meaningful for this dataset.

\vspace{0.07in}
\noindent
\textbf{Baselines.} For HMDB51 and UCF101, we use three  attention-based methods (Visual attention \cite{sharma2015action}, VideoLSTM \cite{li2018videolstm}, and Attentional Pooling \cite{girdhar2017attentional})  and our own method without attention (ResNet101-ImageNet) as baselines. For the Moments in Time dataset, as there are no attention-based methods results available, we use ResNet50-ImageNet \cite{monfort2018moments}, TSN-Spatial \cite{wang2016temporal} and TRN-Multiscale \cite{zhou2017temporal} as our baselines. 

\vspace{0.07in}
\noindent
\textbf{Experimental setup.} We use the same parameters for HMDB51 and UCF101: a single convolutional LSTM layer with 512 hidden-state dimensions, sequence length $n=25$, $\lambda_{\mathrm{TV}}=10^{-5}$, $\lambda_{\mathrm{contrast}}=10^{-4}$, and $\lambda_{\mathrm{unimodal}}=1$.  For the Moments in Time dataset, we use time sequence length $n=15$. The dataset pre-processing and data augmentation are the same as the ResNet ImageNet experiment \cite{he2016deep}. For more details on the experimental setup, please refer to the supplementary material.

\vspace{0.07in}
\noindent
\textbf{Qualitative results.} We visualize the spatial attention and temporal attention results in Fig.~\ref{fig:attention_hmdb51}. The spatial attention can correctly focus on the important spatial areas of the image, and the temporal attention shows a unimodal distribution for the entire action from starting frame to final frame. %It also indicates our spatial attention mechanism is effective input for later temporal attention module. 
More results are shown in the supplementary material. 

\vspace{0.07in}
\noindent
\textbf{Quantitative results.} We show the top-1 video action classification accuracy for HMDB51 and UCF101 datasets in Table \ref{tab:hmdb51}. Our proposed model outperforms previous attention based models \cite{girdhar2017attentional,li2018videolstm,sharma2015action} and ResNet101-ImageNet with the same input.  The ablation experiments demonstrate that all the sub-components of the proposed method contribute to improving the final performance. The results on the Moments in Time dataset are reported in Table \ref{tab:moments_in_time}. Our method achieves the best accuracy comparing to other methods using just RGB modality input. Furthermore, our spatial-temporal attention mechanism is an easy plug-in model which could be based on different architectures. As shown in Table \ref{tab:hmdb51_different_basenetwork}, our spatial-temporal attention mechanism can boost performance for different base networks. It also indicates that base networks are also crucial for the final performance as higher-capacity base networks can provide better image features as input to our model.  
\begin{table} [!hb]
\centering
\begin{tabular}{l|ll}
\hline
\textbf{Model} & \textbf{HMDB51} & \textbf{UCF101}   \\
 \hline 
Visual attention \cite{sharma2015action} &41.31 &84.96 \\
VideoLSTM \cite{li2018videolstm} & 43.30& 79.60\\
Attentional Pooling \cite{girdhar2017attentional}&50.80 &--\\
ResNet101-ImageNet  & 50.04 &83.30\\
\hline
\textbf{Ours}& \textbf{53.07} & \textbf{87.11}  \\ 
\hline
\hline 
\multicolumn{3}{c}{\textbf{Ablation Experiments}} \\
\hline
Ours w/o spatial attention &   51.98 & 85.78\\ 
Ours w/o temporal attention &  52.25& 85.86\\
Ours w/o $L_{\mathrm{TV}}$ & 52.01 & 85.89\\
Ours w/o $L_{\mathrm{contrast}}$ & 52.10 &85.98\\
Ours w/o $L_{\mathrm{unimodal}}$ & 52.05 &86.10\\
\hline 
\end{tabular}
\vspace{+3mm}
\caption{\textbf{Top-1 accuracy (\%) on HMDB51 and UCF101 dataset.} }
\label{tab:hmdb51}
\end{table}

\begin{table}
\centering
\begin{tabular}{l|ll}
\hline
\textbf{Model} & \textbf{Top-1} (\%) & \textbf{Top-5} (\%) \\
 \hline 
ResNet50-ImageNet \cite{monfort2018moments}   & 26.98 &51.74 \\
TSN-Spatial \cite{wang2016temporal}  &24.11 &49.10 \\
TRN-Multiscale \cite{zhou2017temporal}   & 27.20 & 53.05 \\
%\hline 
%BNInception-Flow \cite{monfort2018moments} & Optical flow  &11.60 &27.40 \\
%ResNet50-DyImg \cite{monfort2018moments} & Optical flow & 15.76 &35.69 \\
%TSN-Flow \cite{wang2016temporal} & Optical flow  &15.71 & 34.65 \\
 %\hline 
%TSN-2stream \cite{wang2016temporal} & RGB+Optical flow &25.32 &50.10 \\
%TRN-Multiscale \cite{zhou2017temporal} & RGB+optical flow & \textbf{28.27} & \textbf{53.87}\\
\hline
\textbf{Ours}  & \textbf{27.86} & \textbf{53.52} \\
\hline 
\end{tabular}
\vspace{+3mm}
\caption{\textbf{Results on Moments in Time dataset with RGB modality.} ResNet50-ImageNet and TRN-Multiscale spatial results reported here are based on authors' publicly released trained model.}
\label{tab:moments_in_time}
\end{table}

\begin{table} 
\centering
\begin{tabular}{l|lll}
\hline
\textbf{Base network} & ResNet50 & ResNet101 &ResNet152 \\
\hline
\textbf{w/o attention} & 47.5 & 50.0 & 50.4 \\
\textbf{w attention} & \textbf{49.8} &\textbf{53.1} &\textbf{54.4} \\ 
\hline
\end{tabular}
\vspace{+3mm}
\caption{\textbf{Top-1 accuracy (\%) on HMDB51 of our spatio-temporal attention mechanism with different base networks.} It shows our spatial-temporal attention mechanism can boost performance for different base networks. It also indicates base networks plays a vital role in the final performance as high-capacity base model can provide more accurate image features. }
\label{tab:hmdb51_different_basenetwork}
\end{table}

\subsection{Weakly supervised action localization}
Due to the existence of spatial and temporal attention mechanisms, our model can not only classify the action of the video, but also give a better interpretability of the results, \textit{i.e.} telling which region and frames contribute more to the prediction. In other words, our proposed model can also localize the most discriminant region and frames at the same time. To verify this, we conduct the spatial localization and temporal localization experiments. 

%The spatial localization is to detect the most important region in the image. With the explicit attention mask prediction in our spatial attention module, the region with higher value in the mask is more important naturally. To evaluate the spatial localization result, we first convert the spatial attention map into bounding box, and compare it with the ground truth bounding box.

%The temporal localization is to find the start and end frame of the action in the video. Intuitively, the weight of each frame predicted by the temporal attention module in our model indicates the importance of each frame. By setting a threshold, we can easily get the frame range which are more important to the prediction. We evaluate our temporal localization performance using the mean average precision at different IoU threshold.

\begin{figure}[!ht]
\centering
\begin{minipage}{0.155\textwidth}
\centering
\includegraphics[width=\textwidth]{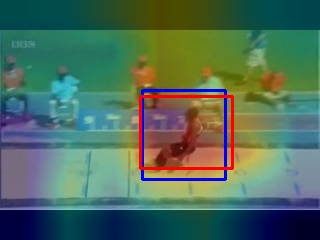}
{(a)}
\end{minipage} 
\begin{minipage}{0.155\textwidth}
\centering
\includegraphics[width=\textwidth]{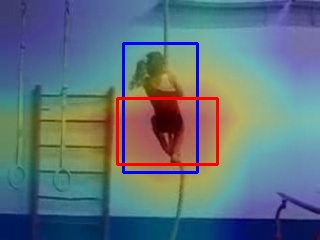}
{(b) }
\end{minipage} 
\begin{minipage}{0.155\textwidth}
\centering
\includegraphics[width=\textwidth]{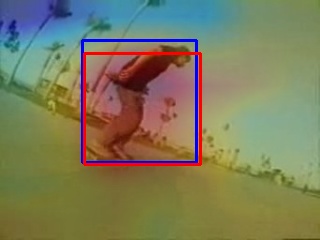}
{(c)}
\end{minipage}
\begin{minipage}{0.155\textwidth}
\centering
\includegraphics[width=\textwidth]{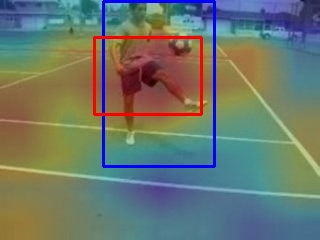}
{(d) }
\end{minipage} 
\begin{minipage}{0.155\textwidth}
\centering
\includegraphics[width=\textwidth]{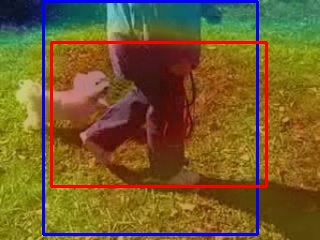}
{(e)}
\end{minipage} 
\begin{minipage}{0.155\textwidth}
\centering
\includegraphics[width=\textwidth]{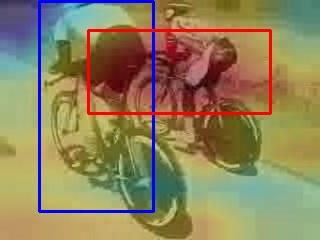}
{(f)}
\end{minipage} 
\vspace{+3mm}
\caption{\textbf{Examples of spatial attention for action localization.} (Best viewed in color.) Blue bounding boxes represent ground truth while the red ones are predictions from our learned spatial attention. (a) long jump, (b) rope climbing, (c) skate boarding, (d) soccer juggling, (e) walking with dog, (f) biking.}
\label{fig:spatial_attention_localization}
\end{figure}

\begin{figure*}[!ht]
\centering
\begin{minipage}{0.135\textwidth}
\centering
\includegraphics[width=\textwidth]{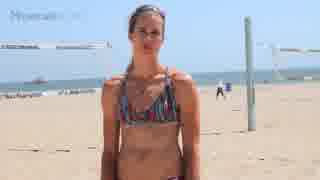}
\end{minipage} 
\vspace{-0.1mm}
\begin{minipage}{0.135\textwidth}
\centering
\includegraphics[width=\textwidth]{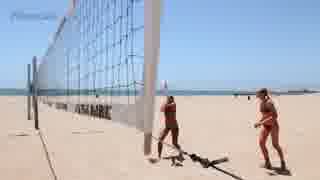}
\end{minipage}
\begin{minipage}{0.135\textwidth}
\centering
\includegraphics[width=\textwidth]{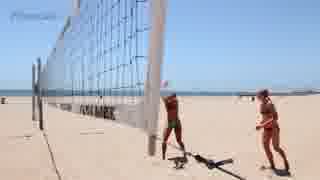}
\end{minipage}
\begin{minipage}{0.135\textwidth}
\centering
\includegraphics[width=\textwidth]{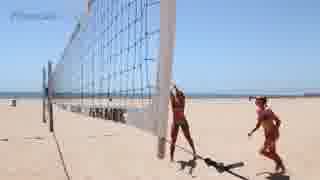}
\end{minipage} 
\begin{minipage}{0.135\textwidth}
\centering
\includegraphics[width=\textwidth]{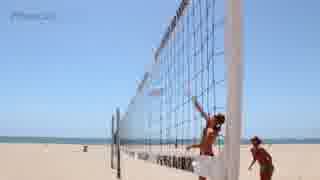}
\end{minipage} 
\begin{minipage}{0.135\textwidth}
\centering
\includegraphics[width=\textwidth]{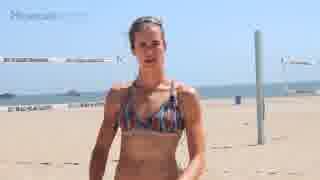}
\end{minipage} 
\begin{minipage}{0.135\textwidth}
\centering
\includegraphics[width=\textwidth]{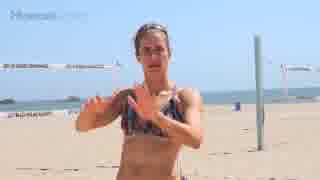}
\end{minipage} 
\begin{minipage}{0.135\textwidth}
\centering
\includegraphics[width=\textwidth]{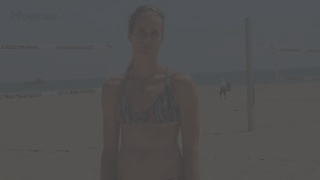}
\end{minipage} 
\vspace{+0.6mm}
\begin{minipage}{0.135\textwidth}
\centering
\includegraphics[width=\textwidth]{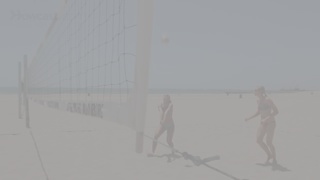}
\end{minipage}
\vspace{+0.6mm}
\begin{minipage}{0.135\textwidth}
\centering
\includegraphics[width=\textwidth]{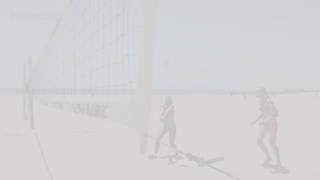}
\end{minipage}
\vspace{+0.6mm}
\begin{minipage}{0.135\textwidth}
\centering
\includegraphics[width=\textwidth]{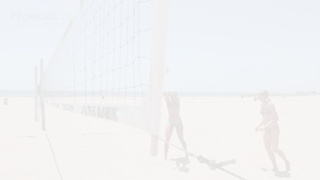}
\end{minipage} 
\vspace{+0.6mm}
\begin{minipage}{0.135\textwidth}
\centering
\includegraphics[width=\textwidth]{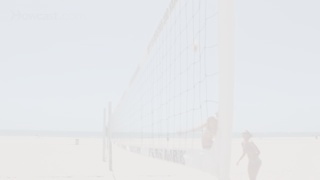}
\end{minipage} 
\vspace{+0.6mm}
\begin{minipage}{0.135\textwidth}
\centering
\includegraphics[width=\textwidth]{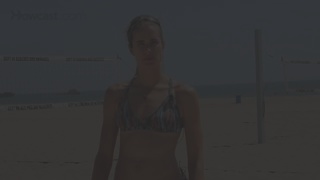}
\end{minipage} 
\vspace{+0.6mm}
\begin{minipage}{0.135\textwidth}
\centering
\includegraphics[width=\textwidth]{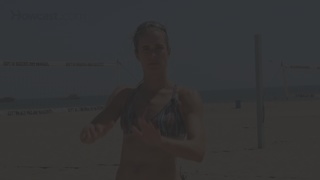}
\end{minipage} 
\vspace{+0.6mm}
\begin{minipage}{0.135\textwidth}
\centering
\includegraphics[width=\textwidth]{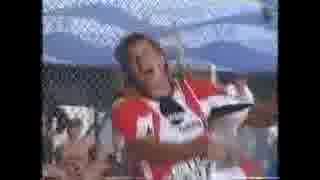}
\end{minipage} 
\begin{minipage}{0.135\textwidth}
\centering
\includegraphics[width=\textwidth]{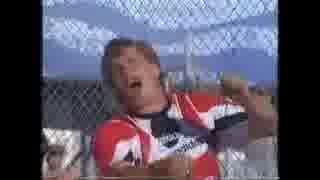}
\end{minipage}
\begin{minipage}{0.135\textwidth}
\centering
\includegraphics[width=\textwidth]{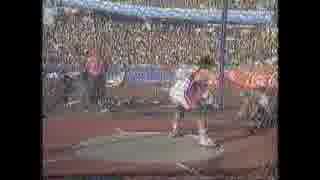}
\end{minipage}
\begin{minipage}{0.135\textwidth}
\centering
\includegraphics[width=\textwidth]{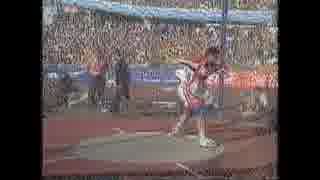}
\end{minipage} 
\begin{minipage}{0.135\textwidth}
\centering
\includegraphics[width=\textwidth]{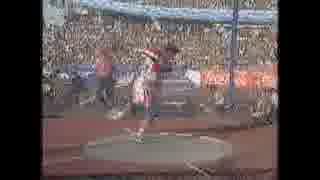}
\end{minipage} 
\begin{minipage}{0.135\textwidth}
\centering
\includegraphics[width=\textwidth]{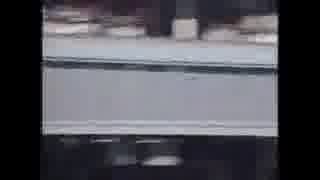}
\end{minipage} 
\begin{minipage}{0.135\textwidth}
\centering
\includegraphics[width=\textwidth]{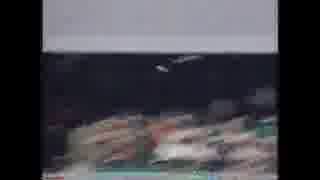}
\end{minipage} 
\begin{minipage}{0.135\textwidth}
\centering
\includegraphics[width=\textwidth]{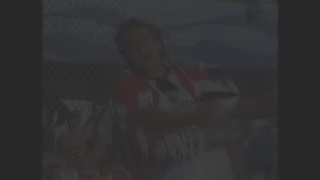}
\end{minipage} 
\begin{minipage}{0.135\textwidth}
\centering
\includegraphics[width=\textwidth]{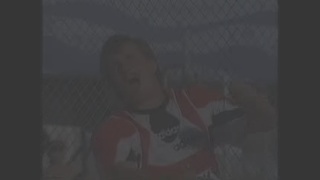}
\end{minipage}
\begin{minipage}{0.135\textwidth}
\centering
\includegraphics[width=\textwidth]{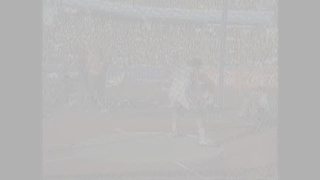}
\end{minipage}
\begin{minipage}{0.135\textwidth}
\centering
\includegraphics[width=\textwidth]{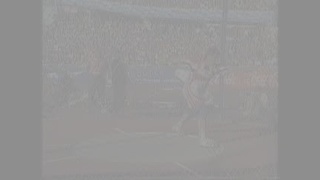}
\end{minipage} 
\begin{minipage}{0.135\textwidth}
\centering
\includegraphics[width=\textwidth]{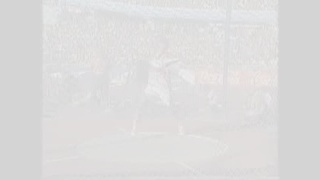}
\end{minipage} 
\begin{minipage}{0.135\textwidth}
\centering
\includegraphics[width=\textwidth]{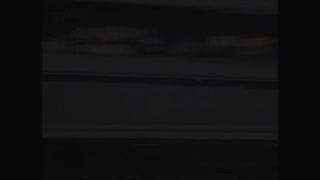}
\end{minipage} 
\begin{minipage}{0.135\textwidth}
\centering
\includegraphics[width=\textwidth]{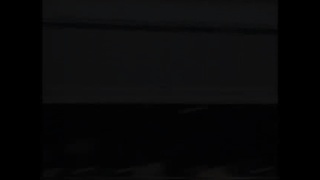}
\end{minipage} 
\vspace{+3mm}
\caption{\textbf{Examples of temporal attention from THUMOS14.} The upper two rows show original images of the \emph{Volleyball} action, and the corresponding images overlaid with temporal attention weights. The lower two rows show the \emph{Throw Discus} action. Our temporal attention module can automatically highlight important frames and avoid irrelevant frames corresponding to non-action poses or background.}
\label{fig:temporal_localization_attention}
\end{figure*}  

\begin{table*}[!hb]
\centering
\begin{tabular}{l|llll}
\hline

\textbf{Methods}&\textbf{$\alpha=0.05$}&\textbf{$\alpha=0.1$}& \textbf{$\alpha=0.2$}& \textbf{$\alpha=0.3$}\\
\hline
\textbf{Fast action proposal$^*$} \cite{yu2015fast} &42.8\% &-- &-- &--\\
\textbf{Learning to track$^*$} \cite{weinzaepfel2015learning} &54.3\% &51.7\%& \textbf{47.7}\%& \textbf{37.8}\% \\
\hline 
%\textbf{Averaging all the bounding boxes} & & \\
\textbf{Ours} & \textbf{67.0}\% &\textbf{58.2}\% &40.2\% &30.7\%\\
\hline 
\end{tabular}
\vspace{+3mm}
\caption{\textbf{Spatial action localization results on UCF101-24 dataset} measured by
mAP at different IoU thresholds $\alpha$. *The baseline methods are strongly supervised spatial localization methods.}
\label{tab:ucf101-24}
\end{table*}

\begin{table*}[!hb]
\centering
\begin{tabular}{l|lllll}
\hline
\textbf{Method}&\textbf{$\alpha=0.1$}&\textbf{$\alpha=0.2$}& \textbf{$\alpha=0.3$}& \textbf{$\alpha=0.4$}&\textbf{$\alpha=0.5$}\\
\hline
REINFORCE \cite{yeung2016end} & 48.9\% &44.0\% & 36.0\% &26.4\% &17.1\% \\
UntrimmedNet \cite{wang2017untrimmednets} & 44.4\%& 37.7\%&28.2\% &21.1\% & 13.7\%\\
\hline 
\textbf{Ours} & \textbf{70.0}\% & \textbf{61.4}\%  &\textbf{48.6}\% &\textbf{32.6}\% & \textbf{17.9}\% \\
\hline 
\end{tabular}
\vspace{+3mm}
\caption{\textbf{Temporal action localization results on THUMOS'14 dataset} measured by mAP at different IoU thresholds $\alpha$. }
\label{tab:thumos_14_20}
\end{table*}

\subsubsection{Spatial action localization}
The spatial localization is designed to detect the most important region in the image. With the explicit attention mask prediction in our spatial attention module, the region with higher value in the mask is considered naturally more important.

\vspace{0.07in}
\noindent
\textbf{Dataset.} UCF101-24 is a subset of 24 classes out of 101 classes of UCF101 \cite{soomro2012ucf101} that comes with spatio-temporal localization annotation,  in the form of bounding box annotations for humans, with THUMOS2013 and THUMOS2014 challenges \cite{THUMOS14}.

\vspace{0.07in}
\noindent
\textbf{Baselines and error metric.} We use supervised methods Fast action proposal \cite{yu2015fast} and Learning to track \cite{weinzaepfel2015learning} as our baselines as currently there are no quantitative results available for weakly supervised setting. Fast action proposal \cite{yu2015fast} is formulated as a maximum set coverage problem and solved by a greedy-based method, and can capture spatial-temporal video tube of high potential to localize human action. Learning to track \cite{weinzaepfel2015learning} first detects frame-level proposals and scores them with a combination of static and motion CNN features. It then tracks high-scoring proposals throughout the video using a tracking-by-detection approach. Note that these supervised spatial localization methods require bounding box annotations, while our method only needs classification labels.

We use Intersection Over Union (IoU) of the predicted and ground truth bounding boxes as our error metric.

\vspace{0.07in}
\noindent
\textbf{Experimental setup.} For training, we only use the classification labels without spatial bounding box annotations. For evaluation, we threshold the produced saliency mask at 0.5 and the tightest bounding box that contains the thresholded saliency map is set as the predicted localization box for each frame. The predicted localization boxes are compared with the ground truth bounding boxes at different IoU levels. 

\vspace{0.07in}
\noindent
\textbf{Qualitative results.} 
We show some qualitative results in Fig.~\ref{fig:spatial_attention_localization}. Our spatial attention can attend to important action areas. The ground truth bounding boxes include all the human actions, while our attention could attend to crucial parts of an action such as in Fig.~\ref{fig:spatial_attention_localization}(d) and (e). Furthermore, our attention mechanism is able to attend to areas with multiple human actions. For instance, in Fig.~\ref{fig:spatial_attention_localization}(f) the ground truth only includes one person bicycling, but our attention can include both people bicycling. More qualitative results including failure cases are included in the supplementary material.

\vspace{0.07in}
\noindent
\textbf{Quantitative results.}  Table \ref{tab:ucf101-24} shows the quantitative results for UCF101-24 spatial localization results. Our attention mechanism works better compared with the baseline methods when the IoU threshold is lower mainly because our model only focuses on important spatial areas rather than the entire human action annotated by bounding boxes. Compared with the baseline methods trained with ground truth bounding boxes, we only use the action classification labels, no ground truth bounding boxes are used.  

\subsubsection{Temporal action localization}
The temporal localization is designed to find the start and end frame of the action in the video. Intuitively, the weight of each frame predicted by the temporal attention module in our model indicates the importance of each frame.

\vspace{0.07in}
\noindent
\textbf{Dataset.} The action detection task of THUMOS14 \cite{THUMOS14} consists
of 20 classes of sports activities, and contains 2765 trimmed videos for training, and 200 and 213 untrimmed videos for validation and test, respectively. Following the standard practice \cite{yeung2016end, SSN2017ICCV}, we use the validation set as training and evaluate on the testing set. To avoid the training ambiguity, we remove the videos with multiple labels. We extract RGB frames from the raw videos at 10 fps.

\vspace{0.07in}
\noindent
\textbf{Baselines and error metric.} We compare our method with a reinforcement learning based method REINFORCE \cite{yeung2016end} and a weakly supervised method UntrimmedNets \cite{wang2017untrimmednets} for temporal action localization. \cite{yeung2016end} formulates the model as a recurrent neural network-based agent that interacts with a video over time, and use REINFORCE to learn the agent's decision policy. UntrimmedNets \cite{wang2017untrimmednets} couple both the classification module and the selection module to learn the action models and reason about the temporal duration of action instances.

Intersection Over Union (IoU) of the predicted and ground truth temporal labels are used as our error metric.

\vspace{0.07in}
\noindent
\textbf{Experimental setup.} We use the same hyperparameters for THUMOS14 as HMDB51, UCF101 and UCF101-24. For training, we only use the classification labels without temporal annotation labels. For evaluation, we threshold the normalized temporal attention importance weight at 0.5. These predicted temporal localization frames are then compared with the ground truth annotation at different IoU thresholds. 

\vspace{0.07in}
\noindent
\textbf{Qualitative results.}
We first visualize some examples of learned attention weights on the test data of THUMOS14 in Fig. \ref{fig:temporal_localization_attention}.  We  see  that  our temporal attention module  is  able  to
automatically  highlight  important  frames  and  to  avoid  irrelevant frames corresponding to background or non-action human poses. More qualitative results are included in the supplementary material.

\vspace{0.07in}
\noindent
 \textbf{Quantitative results.} With our spatial temporal attention mechanism, the video action classification accuracy for the THUMOS'14 20 classes improved from $74.45\%$ to $78.33\%$: a $3.88\%$ increase. Besides improving the classification accuracy, we show our temporal attention mechanism is able to highlight discriminative frames quantitatively in Table \ref{tab:thumos_14_20}. Compared with reinforcement learning based method \cite{yeung2016end} and weakly supervised method \cite{wang2017untrimmednets}, our method achieves the best accuracy in terms of different levels of IoU thresholds.  

\section{Conclusion}
In this paper we have introduced an interpretable spatial-temporal attention mechanism for video action recognition. Also, we propose  a  set  of  regularizers that ensure our attention mechanism attends to coherent regions in space and time, further improving the performance. Besides boosting video action recognition accuracy, our easy-plug-in spatio-temporal mechanism can increase the model interpretability. Furthermore, we qualitatively and quantitatively demonstrate that our spatio-temporal attention is able to localize discriminative regions and important frames, despite being trained in a purely weakly-supervised manner with only classification labels. Experimental results demonstrate the efficacy of our approach, showing superior or comparable accuracy with the state-of-the-art methods. For future work, it will be promising to integrate our spatio-temporal attention mechanism into 3D ConvNets \cite{carreira2017quo,  hara2018can, tran2018closer} for improving action recognition performance and increasing model interpretability.

{\small
\bibliographystyle{ieee}
\bibliography{egbib}

\begin{thebibliography}{10}\itemsep=-1pt

\bibitem{bahdanau2014neural}
D.~Bahdanau, K.~Cho, and Y.~Bengio.
\newblock Neural machine translation by jointly learning to align and
  translate.
\newblock In {\em ICLR}, 2015.

\bibitem{carreira2017quo}
J.~Carreira and A.~Zisserman.
\newblock Quo vadis, action recognition? a new model and the kinetics dataset.
\newblock In {\em CVPR}, 2017.

\bibitem{deng2009imagenet}
J.~Deng, W.~Dong, R.~Socher, L.-J. Li, K.~Li, and L.~Fei-Fei.
\newblock Imagenet: A large-scale hierarchical image database.
\newblock In {\em CVPR}, 2009.

\bibitem{du2018recurrent}
W.~Du, Y.~Wang, and Y.~Qiao.
\newblock Recurrent spatial-temporal attention network for action recognition
  in videos.
\newblock {\em IEEE Transactions on Image Processing (ICIP)}, 2018.

\bibitem{feichtenhofer2016convolutional}
C.~Feichtenhofer, A.~Pinz, and A.~Zisserman.
\newblock Convolutional two-stream network fusion for video action recognition.
\newblock In {\em CVPR}, 2016.

\bibitem{girdhar2017attentional}
R.~Girdhar and D.~Ramanan.
\newblock Attentional pooling for action recognition.
\newblock In {\em NIPS}, 2017.

\bibitem{hara2018can}
K.~Hara, H.~Kataoka, and Y.~Satoh.
\newblock Can spatiotemporal 3d cnns retrace the history of 2d cnns and
  imagenet?
\newblock In {\em CVPR}, 2018.

\bibitem{he2016deep}
K.~He, X.~Zhang, S.~Ren, and J.~Sun.
\newblock Deep residual learning for image recognition.
\newblock In {\em CVPR}, 2016.

\bibitem{hu2017squeeze}
J.~Hu, L.~Shen, and G.~Sun.
\newblock Squeeze-and-excitation networks.
\newblock In {\em CVPR}, 2018.

\bibitem{huang2017densely}
G.~Huang, Z.~Liu, L.~Van Der~Maaten, and K.~Q. Weinberger.
\newblock Densely connected convolutional networks.
\newblock In {\em CVPR}, 2017.

\bibitem{ioffe2015batch}
S.~Ioffe and C.~Szegedy.
\newblock Batch normalization: Accelerating deep network training by reducing
  internal covariate shift.
\newblock {\em arXiv preprint arXiv:1502.03167}, 2015.

\bibitem{jetley2018learn}
S.~Jetley, N.~A. Lord, N.~Lee, and P.~H. Torr.
\newblock Learn to pay attention.
\newblock {\em ICLR}, 2018.

\bibitem{THUMOS14}
Y.-G. Jiang, J.~Liu, A.~Roshan~Zamir, G.~Toderici, I.~Laptev, M.~Shah, and
  R.~Sukthankar.
\newblock {THUMOS} challenge: Action recognition with a large number of
  classes.
\newblock \url{http://crcv.ucf.edu/THUMOS14/}, 2014.

\bibitem{kim2017interpretable}
J.~Kim and J.~Canny.
\newblock Interpretable learning for self-driving cars by visualizing causal
  attention.
\newblock In {\em CVPR}, 2017.

\bibitem{kong2018human}
Y.~Kong and Y.~Fu.
\newblock Human action recognition and prediction: A survey.
\newblock {\em arXiv preprint arXiv:1806.11230}, 2018.

\bibitem{kuehne2011hmdb}
H.~Kuehne, H.~Jhuang, E.~Garrote, T.~Poggio, and T.~Serre.
\newblock Hmdb: a large video database for human motion recognition.
\newblock In {\em ICCV}, 2011.

\bibitem{li2018tell}
K.~Li, Z.~Wu, K.-C. Peng, J.~Ernst, and Y.~Fu.
\newblock Tell me where to look: Guided attention inference network.
\newblock {\em CVPR}, 2018.

\bibitem{li2018videolstm}
Z.~Li, K.~Gavrilyuk, E.~Gavves, M.~Jain, and C.~G. Snoek.
\newblock Videolstm convolves, attends and flows for action recognition.
\newblock {\em Computer Vision and Image Understanding (CVIU)}, 2018.

\bibitem{lu2016hierarchical}
J.~Lu, J.~Yang, D.~Batra, and D.~Parikh.
\newblock Hierarchical question-image co-attention for visual question
  answering.
\newblock In {\em NIPS}, 2016.

\bibitem{mahendran2016salient}
A.~Mahendran and A.~Vedaldi.
\newblock Salient deconvolutional networks.
\newblock In {\em ECCV}, 2016.

\bibitem{monfort2018moments}
M.~Monfort, B.~Zhou, S.~A. Bargal, A.~Andonian, T.~Yan, K.~Ramakrishnan,
  L.~Brown, Q.~Fan, D.~Gutfruend, C.~Vondrick, et~al.
\newblock Moments in time dataset: one million videos for event understanding.
\newblock {\em arXiv preprint arXiv:1801.03150}, 2018.

\bibitem{ramprasaath2016grad}
R.~Ramprasaath, D.~Abhishek, V.~Ramakrishna, C.~Michael, P.~Devi, and B.~Dhruv.
\newblock Grad-cam: Why did you say that? visual explanations from deep
  networks via gradient-based localization.
\newblock {\em CVPR}, 2016.

\bibitem{rensink2000dynamic}
R.~A. Rensink.
\newblock The dynamic representation of scenes.
\newblock {\em Visual cognition}, 7(1-3):17--42, 2000.

\bibitem{rudin1992nonlinear}
L.~I. Rudin, S.~Osher, and E.~Fatemi.
\newblock Nonlinear total variation based noise removal algorithms.
\newblock {\em Physica D: nonlinear phenomena}, 1992.

\bibitem{selvaraju2017grad}
R.~R. Selvaraju, M.~Cogswell, A.~Das, R.~Vedantam, D.~Parikh, and D.~Batra.
\newblock Grad-cam: Visual explanations from deep networks via gradient-based
  localization.
\newblock In {\em ICCV}, 2017.

\bibitem{sharma2015action}
S.~Sharma, R.~Kiros, and R.~Salakhutdinov.
\newblock Action recognition using visual attention.
\newblock {\em arXiv preprint arXiv:1511.04119}, 2015.

\bibitem{xingjian2015convolutional}
X.~Shi, Z.~Chen, H.~Wang, D.-Y. Yeung, W.-K. Wong, and W.-c. Woo.
\newblock Convolutional lstm network: A machine learning approach for
  precipitation nowcasting.
\newblock In {\em NIPS}, 2015.

\bibitem{shou2016action}
Z.~Shou, D.~Wang, and S.~Chang.
\newblock Action temporal localization in untrimmed videos via multi-stage
  cnns.
\newblock In {\em CVPR}, 2016.

\bibitem{simonyan2013deep}
K.~Simonyan, A.~Vedaldi, and A.~Zisserman.
\newblock Deep inside convolutional networks: Visualising image classification
  models and saliency maps.
\newblock {\em arXiv preprint arXiv:1312.6034}, 2013.

\bibitem{simonyan2014two}
K.~Simonyan and A.~Zisserman.
\newblock Two-stream convolutional networks for action recognition in videos.
\newblock In {\em NIPS}, 2014.

\bibitem{singh2017online}
G.~Singh, S.~Saha, M.~Sapienza, P.~H. Torr, and F.~Cuzzolin.
\newblock Online real-time multiple spatiotemporal action localisation and
  prediction.
\newblock In {\em ICCV}, 2017.

\bibitem{song2017end}
S.~Song, C.~Lan, J.~Xing, W.~Zeng, and J.~Liu.
\newblock An end-to-end spatio-temporal attention model for human action
  recognition from skeleton data.
\newblock In {\em AAAI}, 2017.

\bibitem{soomro2012ucf101}
K.~Soomro, A.~R. Zamir, and M.~Shah.
\newblock Ucf101: A dataset of 101 human actions classes from videos in the
  wild.
\newblock {\em arXiv preprint arXiv:1212.0402}, 2012.

\bibitem{springenberg2014striving}
J.~T. Springenberg, A.~Dosovitskiy, T.~Brox, and M.~Riedmiller.
\newblock Striving for simplicity: The all convolutional net.
\newblock {\em arXiv preprint arXiv:1412.6806}, 2014.

\bibitem{torabi2017action}
A.~Torabi and L.~Sigal.
\newblock Action classification and highlighting in videos.
\newblock {\em arXiv preprint arXiv:1708.09522}, 2017.

\bibitem{tran2015learning}
D.~Tran, L.~Bourdev, R.~Fergus, L.~Torresani, and M.~Paluri.
\newblock Learning spatiotemporal features with 3d convolutional networks.
\newblock In {\em CVPR}, 2015.

\bibitem{tran2018closer}
D.~Tran, H.~Wang, L.~Torresani, J.~Ray, Y.~LeCun, and M.~Paluri.
\newblock A closer look at spatiotemporal convolutions for action recognition.
\newblock In {\em CVPR}, 2018.

\bibitem{vaswani2017attention}
A.~Vaswani, N.~Shazeer, N.~Parmar, J.~Uszkoreit, L.~Jones, A.~N. Gomez,
  {\L}.~Kaiser, and I.~Polosukhin.
\newblock Attention is all you need.
\newblock In {\em NIPS}, 2017.

\bibitem{wang2017untrimmednets}
L.~Wang, Y.~Xiong, D.~Lin, and L.~Van~Gool.
\newblock Untrimmednets for weakly supervised action recognition and detection.
\newblock In {\em CVPR}, 2017.

\bibitem{wang2016temporal}
L.~Wang, Y.~Xiong, Z.~Wang, Y.~Qiao, D.~Lin, X.~Tang, and L.~Van~Gool.
\newblock Temporal segment networks: Towards good practices for deep action
  recognition.
\newblock In {\em ECCV}, 2016.

\bibitem{wang2016hierarchical}
Y.~Wang, S.~Wang, J.~Tang, N.~O'Hare, Y.~Chang, and B.~Li.
\newblock Hierarchical attention network for action recognition in videos.
\newblock {\em arXiv preprint arXiv:1607.06416}, 2016.

\bibitem{weinzaepfel2015learning}
P.~Weinzaepfel, Z.~Harchaoui, and C.~Schmid.
\newblock Learning to track for spatio-temporal action localization.
\newblock In {\em CVPR}, 2015.

\bibitem{xiong2016dynamic}
C.~Xiong, S.~Merity, and R.~Socher.
\newblock Dynamic memory networks for visual and textual question answering.
\newblock In {\em ICML}, 2016.

\bibitem{xu2017r}
H.~Xu, A.~Das, and K.~Saenko.
\newblock {R-C3D}: region convolutional 3d network for temporal activity
  detection.
\newblock In {\em ICCV}, 2017.

\bibitem{xu2016ask}
H.~Xu and K.~Saenko.
\newblock Ask, attend and answer: Exploring question-guided spatial attention
  for visual question answering.
\newblock In {\em ECCV}, 2016.

\bibitem{yeung2016end}
S.~Yeung, O.~Russakovsky, G.~Mori, and L.~Fei-Fei.
\newblock End-to-end learning of action detection from frame glimpses in
  videos.
\newblock In {\em CVPR}, 2016.

\bibitem{yu2015fast}
G.~Yu and J.~Yuan.
\newblock Fast action proposals for human action detection and search.
\newblock In {\em CVPR}, 2015.

\bibitem{zeiler2014visualizing}
M.~D. Zeiler and R.~Fergus.
\newblock Visualizing and understanding convolutional networks.
\newblock In {\em ECCV}, 2014.

\bibitem{zhang2016top}
J.~Zhang, Z.~Lin, J.~Brandt, X.~Shen, and S.~Sclaroff.
\newblock Top-down neural attention by excitation backprop.
\newblock In {\em ECCV}, 2016.

\bibitem{zhang2018interpretable}
Q.~Zhang, Y.~Nian~Wu, and S.-C. Zhu.
\newblock Interpretable convolutional neural networks.
\newblock In {\em CVPR}, 2018.

\bibitem{zhangetal2017mdnet}
Z.~Zhang, Y.~Xie, F.~Xing, M.~McGough, and L.~Yang.
\newblock {MDNet}: A semantically and visually interpretable medical image
  diagnosis network.
\newblock In {\em CVPR}, 2017.

\bibitem{SSN2017ICCV}
Y.~Zhao, Y.~Xiong, L.~Wang, Z.~Wu, X.~Tang, and D.~Lin.
\newblock Temporal action detection with structured segment networks.
\newblock In {\em ICCV}, 2017.

\bibitem{zhou2017temporal}
B.~Zhou, A.~Andonian, and A.~Torralba.
\newblock Temporal relational reasoning in videos.
\newblock In {\em ECCV}, 2018.

\bibitem{zhou2016learning}
B.~Zhou, A.~Khosla, A.~Lapedriza, A.~Oliva, and A.~Torralba.
\newblock Learning deep features for discriminative localization.
\newblock In {\em CVPR}, 2016.

\end{thebibliography}
}

\end{document}